\documentclass[sigconf,screen,authorversion,nonacm]{acmart}

\usepackage{fancyhdr}
\AtBeginDocument{%
    \addtolength{\footskip}{2.0\baselineskip}%
    \fancyfoot[L]{\textit{\textbf{Preprint.}}}%
}

\AtBeginDocument{%
  }

\setcopyright{acmlicensed}
\copyrightyear{2018}
\acmYear{2018}
\acmDOI{XXXXXXX.XXXXXXX}
\acmConference[Conference acronym 'XX]{Make sure to enter the correct
  conference title from your rights confirmation email}{June 03--05,
  2018}{Woodstock, NY}
\acmISBN{978-1-4503-XXXX-X/2018/06}

\usepackage[capitalize]{cleveref}
\usepackage{tikz}
\usepackage{siunitx}
\usepackage{physics}
\usepackage{algorithm}
\usepackage{algpseudocode}
\usepackage{graphicx}
\usepackage{booktabs}

\usepackage{amsmath}
\DeclareFontFamily{U}{stix2bb}{}
\DeclareFontShape{U}{stix2bb}{m}{n} {<-> stix2-mathbb}{}

\NewDocumentCommand{\indicator}{}{\text{\usefont{U}{stix2bb}{m}{n}1}}

\usepackage{math}

\usepackage{xspace}
\makeatletter
\DeclareRobustCommand\onedot{\futurelet\@let@token\@onedot}
\def\@onedot{\ifx\@let@token.\else.\null\fi\xspace}

\def\eg{\emph{e.g}\onedot} 
\def\ie{\emph{i.e}\onedot}

\def\iid{\emph{i.i.d.}\xspace}
\makeatother

\begin{document}

\title{A probabilistic framework for dynamic quantization}

\author{{Gabriele Santini, Francesco Paissan, Elisabetta Farella}}
\email{{gsantini@fbk.eu,   francescopaissan@gmail.com,    efarella@fbk.eu}}
% \author{Francesco Paissan}
% \email{francescopaissan@gmail.com}
% \author{Elisabetta Farella}
% \email{efarella@fbk.eu}
\affiliation{%
  \institution{Fondazione Bruno Kessler}
  \city{Trento}
  % \state{Italy}
  \country{Italy}
}

% \renewcommand{\shortauthors}{Santini, Paissan, Farella}

\begin{abstract}
We propose a probabilistic framework for dynamic quantization of neural networks that allows for a computationally efficient input-adaptive rescaling of the quantization parameters.
%Our framework integrates probabilistic modelling of the neural network's pre-activations using a surrogate model to enable to adaptively adapt the quantization parameters for each input without incurring a significant memory overhead.
Our framework applies a probabilistic model to the network's pre-activations through a lightweight surrogate, enabling the adaptive adjustment of the quantization parameters on a per-input basis without significant memory overhead.
We validate our approach on a set of popular computer vision tasks and models, observing only a negligible loss in performance. Our method strikes the best performance and computational overhead tradeoff compared to standard quantization strategies.
\end{abstract}    




%% the work being presented. Separate the keywords with commas..


\newcommand{\fp}[1]{\textcolor{teal}{FP: #1}}
\newcommand{\ef}[1]{\textcolor{red}{EF: #1}}
\newcommand{\gs}[1]{\textcolor{orange}{GS: #1}}

\newcommand{\Var}[1]{\operatorname{Var}\left[#1\right]}
\newcommand{\Exp}[1]{\E\left[#1\right]}

\maketitle

\section{Introduction} \label{sec:intro}

Neural networks have made remarkable progress across a wide range of applications \cite{touvron2023llama,Rombach2021HighResolutionIS,Radford2022RobustSR}, with larger models consistently outperforming smaller ones in a predictable way \cite{bahri2024explaining, alabdulmohsin2022revisiting}. The merit of this success is shared between new model architectures \cite{Zhu2024VisionME,Liu2021SwinTH,Vaswani2017AttentionIA}, training strategies \cite{Yu2022CoCaCC,Liu2023ImprovedBW} and system optimizations that allow highly over-parametrized models to run efficiently \cite{Bolya2022TokenMY,daoflashattention} even on consumer-grade GP/GPUs. In this paper, we analyze neural network \textit{quantization} \cite{Nagel2021AWP,krishnamoorthi2018quantizing}, one of the earliest and most effective compression techniques that has consistently shown good performance and remains highly relevant given the increasing scale at which neural architectures are operating \cite{Yu2022CoCaCC}.
Quantization is not only crucial to enable inference of large models but also for low-footprint neural architectures \cite{Paissan2021PhiNetsAS,Ancilotto2023XiNetEN,lin2020mcunet,li2021micronet,liu2023efficientvit}, which typically run on devices whose parameter memory and working memory are in the order of few megabytes.

Many quantization schemes have been proposed \cite{krishnamoorthi2018quantizing,Nagel2021AWP}, each offering different trade-offs in performance and computational costs. We can categorize them based on (i) how the quantization mapping is defined - symmetric and asymmetric quantization - and (ii) how the quantization parameters are computed - static and dynamic quantization. This paper will focus on \emph{uniform affine quantization}, also known as asymmetric quantization. This quantization scheme, described in detail in \cref{sec:model_q}, is the most widely used as it permits efficient implementation of fixed point arithmetic \cite{Nagel2021AWP} on a variety of hardware platforms without introducing a significant quantization error. Additionally, it is worth noting that symmetric quantization is a specific instance of uniform affine quantization; therefore, the results derived in this paper also apply to symmetric quantization.

The differences between static and dynamic quantization mainly concern how the quantization parameters are estimated. On a first approximation, we can say that the quantization parameters depend only on the dynamic range of the quantity (or tensor) to be quantized. In static quantization, we assume that it is possible to accurately model the dynamic range of the pre-activations of each layer in the neural network by observing them on a given subset of the training data \cite{Nagel2021AWP}. This assumption is inaccurate for common neural network application scenarios, as the model lifetime is shifting to a two-stage pipeline, where pre-training, post-training, and inference happen on distinct data distributions \cite{kumar2025llm}. Other common scenarios where this assumption breaks are zero-shot inference, where the data distribution is unknown beforehand \cite{paissan2024tinyclap,radford2021learning}, and shifts related to ambient noise. To address this issue, dynamic quantization measures the dynamic range of each pre-activation tensor in the network \cite{liu2022instance} for every input. It is, therefore, more robust to shifts in the input data distribution, which would also be propagated to the pre-activations. This also makes dynamic quantization the go-to choice for sequence modelling problems and architectures \cite{vaswani2017attention, hochreiter1997long}. Notably, this flexibility comes at a higher computational overhead, as described in \cref{sec:model_q}.

In this paper, we set out to integrate the computational benefits of static quantization with the performance of dynamic quantization by proposing a novel, probabilistic quantization scheme. Our method works by estimating the dynamic range of the output using a surrogate model of the neural network's pre-activations, enabling input-adaptive quantization with minimal memory overhead. The contributions of this paper are as follows:
\begin{itemize}
    \item we propose a framework to reduce the memory overhead of dynamic quantization while controlling the latency with a scaling parameter, $\gamma$;
    \item we benchmark the performance of our method on five vision tasks at different model scales;
    \item we study the performance drop of different quantization strategies to domain shifts; % to domain shifts demonstrate the effectiveness of our quantization scheme on shifts of the data distribution;
    \item we implement our novel quantization strategy within CMSIS-NN \cite{lai2018cmsis} and benchmark it on an STM32L476RG development board.
\end{itemize}

In the following, \cref{sec:bkg} introduces the mathematical notation used throughout the manuscript and the related works. \cref{sec:model_q} describes the computational costs of the static and dynamic quantization schemes. \cref{sec:methods} presents our framework and its computational complexity. Finally, \cref{sec:expdesign,sec:results} present the experimental setup employed in this paper and the results. The code to reproduce all results is made available for both PyTorch and Embedded C for the on-device experiments.\footnote{\href{https://github.com/Sentz98/ProbabilisticDynamicQuantization}{https://github.com/Sentz98/ProbabilisticDynamicQuantization}}.

% \begin{itemize}
%     \item Quantization is important to allow bigger models to run on resource-constrained device
%     \item Many type of q, each with different perf/compl trade-offs (static, dynamic)
%     \begin{itemize}
%         \item For some models, static doesn't work 
%     \end{itemize}
%     \item We develop a framework to reduce the overhead (const mem footprint, op count) of dynamic quantization in a controllable way. This can be embedded into QAT pipelines.
% \end{itemize}

\begin{figure*}[t]
    \centering
    \resizebox{0.99\linewidth}{!}{
        \begin{tikzpicture}
            \node(img1) {
                \includegraphics[width=0.33\linewidth, trim={0 0 11.5cm 0}]{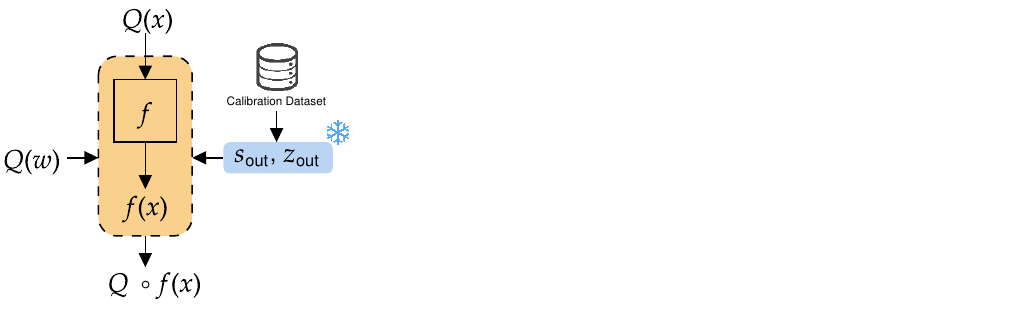}
            };
            \node[right of=img1, xshift=5cm] (img2) {
                \includegraphics[width=0.33\linewidth, trim={0 0 11.5cm 0}]{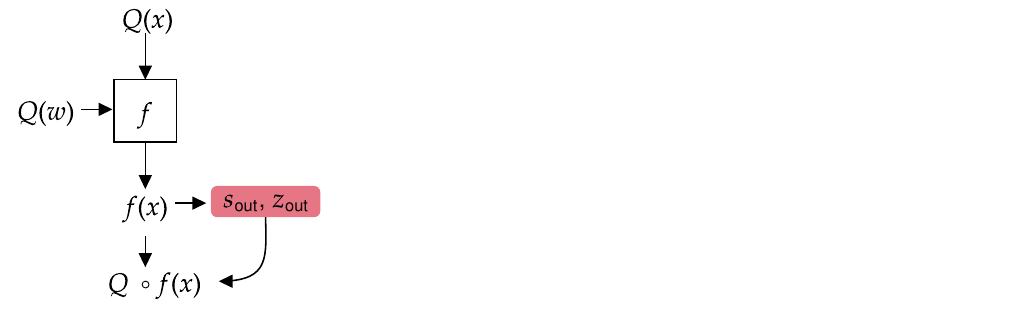}
            };
            \node[right of=img2, xshift=3.5cm] (img3) {
                \includegraphics[width=0.33\linewidth, trim={0 0 11.5cm 0}]{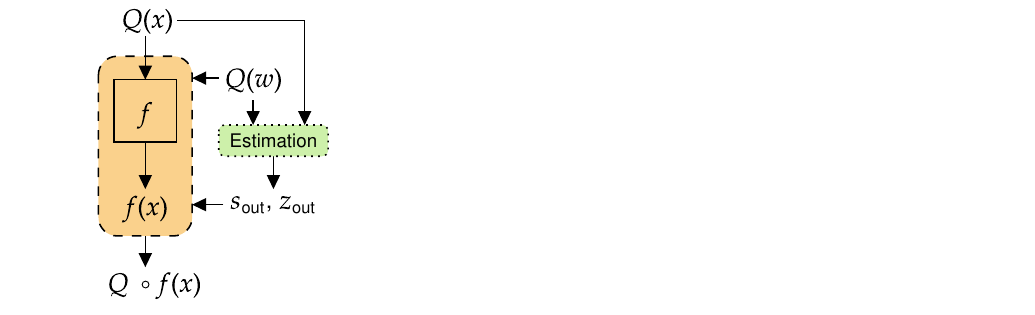}
            };

            % Subfigure notes
            \node [below of=img1, yshift=-2cm] {a)};
            \node [below of=img2, yshift=-2cm] {b)};
            \node [below of=img3, yshift=-2cm] {c)};
        \end{tikzpicture}
    }
    \caption{This diagram showcases the differences between static (a), dynamic (b) and our quantization strategies (c) for a function $f(x, w)$. The quantization parameters for $w$ are always computed before model deployment. For $x$, it depends on the quantization strategy, as explained in \cref{sec:model_q}. The orange box indicates that the output of $f$ is already quantized using the provided parameters. Therefore, $f(x)$ is \emph{never} stored as \SI{32}{\bit} variable. The blue box indicates that the parameters are computed \textit{before model deployment}. The red box indicates that the parameters are computed \textit{after} each evaluation of the function of interest. The green box indicates that the parameters are computed \textit{before} evaluating the function $f$.}
    \label{fig:quant}
\end{figure*}
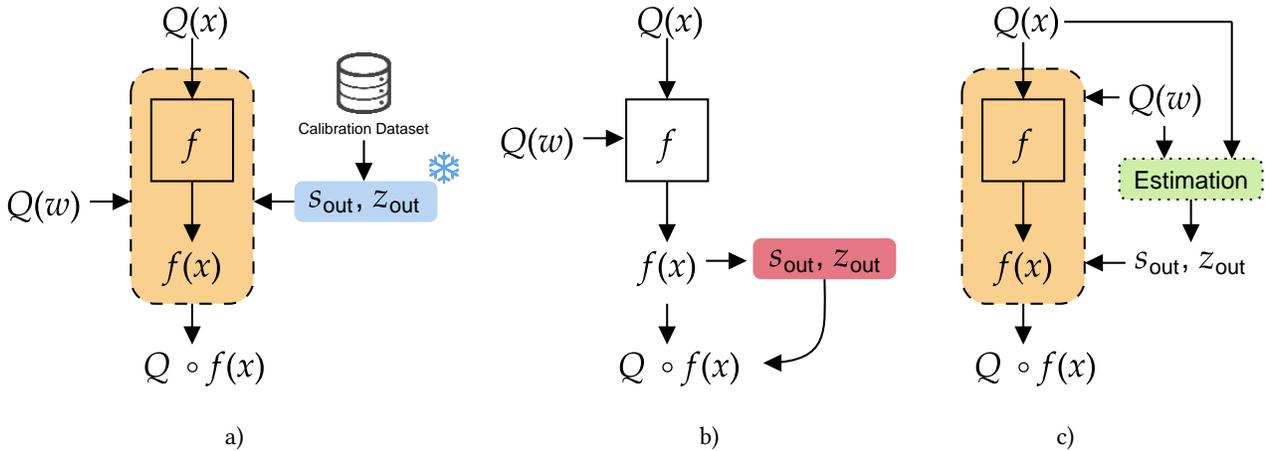
\section{Background} \label{sec:bkg}
In \cref{subsec:notation}, we present the mathematical notation used in the remainder of the paper. \cref{subsec:related} presents recent literature on neural network quantization.

\subsection{Notation} \label{subsec:notation}
Quantization is defined using three parameters: \textit{scale} $s \in \mathbb{R}$, \textit{zero-point} $z \in \mathbb{Z}_\text{b}$ and \textit{bit-width} $b \in \mathbb{N}$. $\mathbb{Z}_\text{b}$ denotes the quantization grid, \ie the set of integers that can be represented with $b$ \si{\bit}.
The scale and zero-point are used to map floating point numbers onto the integer grid, whose width depends on the bit-width. This mapping is described as follows:
\begin{align} \label{eq:quant}
    Q_b(x, s, z) = \text{clamp}\left(\left\lfloor\frac{x}{s}\right\rceil + z; 0, 2^b-1 \right),
\end{align}
where $\left\lfloor\cdot\right\rceil$ is the rounding operation that maps floating point values to the closest integer on the quantization grid
% , $\mathbb{Z}^\text{(b)}$, which is defined by the quantized values data type, $\mathbb{Z}^\text{(b)} = \mathbb{Z}^\text{(b)} \cap [-2^{b-1}, 2^{b-1} - 1]$ for signed data types. 
and the $\text{clamp}(\cdot)$ function is defined as

\begin{align}
    \text{clamp}(x; a, b) =
    \begin{cases}
        a, \qquad x < a, \\
        x, \qquad  a < x < b, \\
        b, \qquad x > b.
    \end{cases}
\end{align}

In the remainder of the paper, when we refer to a quantized variable with $Q_b(x)$, we assume the scale and zero-point are computed by analyzing the maximum ($M$) and minimum ($m$) observed values for the variable to be quantized
\begin{align} \label{eq:qparams}
    s = \frac{M-m}{2^b-1}, \qquad z = -\left\lfloor\frac{m}{s}\right\rceil - 2^{b-1}.
\end{align}

% As any many-to-one mapping, quantization is a lossy compression technique.
We can approximately \emph{de-quantize} a variable using
\begin{align}
    x \approx \hat{x} = s(Q_b(x) - z).
\end{align}

% Note that the de-quantization process is relevant for neural network inference, as it is used during the evaluation of algebraic operators to avoid numeric overflow errors, as detailed in \cref{subsec:cost}.

On vectors and tensors, the quantization operation differs based on how many quantization parameters are used. In \emph{per-tensor quantization}, the quantization parameters are shared among an entire tensor, and quantization is performed element-wise. Another commonly used mode is the \emph{per-channel quantization}, where each channel has a different set of quantization parameters, and the quantization process runs independently on different channels.

\subsection{Related works} \label{subsec:related}

The central research question in quantization is \emph{how can we train neural networks that are robust to compression?} To address this problem, many works in the quantization literature are divided between optimizing quantization schemes and providing novel training strategies to increase the robustness to quantization noise. Hereafter, we present an overview of relevant arts.

\noindent \textbf{Quantization schemes.} The most commonly used quantization scheme in quantization is static quantization \cite{Nagel2021AWP,krishnamoorthi2018quantizing}. As mentioned in \cref{sec:intro}, this can harm the neural network performance significantly due to unexpected shifts in the data distribution. For this reason, when modelling sequences or targeting broad application domains, dynamic quantization is the go-to solution, as the quantization parameters (\cref{eq:qparams}) are defined for each input independently. 
 Noting that the different layers in neural networks have different sensitivity to quantization, we can adjust the bit-width of each layer achieving what is called \textit{mixed precision} quantization \cite{bablani2023efficient}. Many papers proposed different strategies to set the bit-width of each layer, from evolutionary algorithms \cite{Liu2021EvolutionaryQO} to gradient-based optimization \cite{defossezdifferentiable}.
\citet{dong2019hawq,dong2020hawq} propose to set the bit-width of each layer by analyzing the hessian matrix of the weights. \citet{liu2022instance} take this approach one step further by adapting each layers' bit-width for each input, noting that data quality has a role in determining the sensitivity of the neural networks to quantization. Similarly, \citet{Wang2023DataQM} achieves the same objective using a reinforcement learning approach.

\noindent \textbf{Joint Optimization.} To make neural networks more robust to quantization noise, one common strategy is to jointly adapt the quantization scheme and the model's parameters using \textit{quantization-aware training}. Originally, \citet{hubara2018quantized} proposed to train neural networks simulating a quantized forward step. During backpropagation, instead, the gradient of the non-differentiable \cref{eq:quant} is estimated using a straight-through approximation \cite{bengio2013estimating}.
\citet{hubara2018quantized,Nagel2021AWP,krishnamoorthi2018quantizing} describe the impact and tradeoffs of training neural networks and the straight-through approximation used to train neural networks. Notably, in basic quantization-aware training, the quantization parameters are still estimated using \cref{eq:qparams}. In \citet{esser2020learned}, the authors propose to learn the quantization scale ($s$) using gradient-based optimization. This training strategy resembles static quantization with learnable bit-widths. \citet{defossezdifferentiable} take the gradient-based optimization one step further, enabling gradient-based optimization of the bit-width during neural network training. This is achieved using a stochastic quantizer that models the quantization step as differentiable probabilistic mapping. \citet{nikolic2024bitpruning} achieve the same result by approximating the quantization step as a linear interpolation of the two closest integers to the quantized value.

To the best of our knowledge, we are the first to work towards making the dynamic quantization process more efficient for neural networks. We note that our quantization scheme can work alongside any of the state-of-the-art quantization techniques described above. Our approach should be considered a plug-and-play optimization to trade working memory with a scalable computation overhead. This is crucial when performing dynamically quantized inference on devices with limited memory, or neural networks with huge compute requirements.
% Optimization-based: 
% esser2020learned -- symmetric only. learns the quant scale via gradient descent. 
% nikolic2024bitpruning -- very good. can be integrated with ours. only learns bitwidths using a dynamic quantization scheme.
% defossezdifferentiable -- similar to nikolic2024bitpruning, uses a stochastic quantizer

% Mixed Precision:
% dong2020hawq, dong2019hawq -- hessian matrix analysis to extract bitwidth
% liu2022instance -- dyn q. variable bitwidth at inference FOR EACH input
% Wang2023DataQM -- similar to liu2022instance but uses RL
% Liu2021EvolutionaryQO -- evolutionary algorithms to learn the bitwidths

% Inference frameworks (maybe describe some impl details)?
% \cite{lai2018cmsis} \cite{david2021tensorflow}
\section{Analytical Model of Quantization} \label{sec:model_q}

% In this Section, we introduce the mathematical notation and fundamental concepts needed to understand the remainder of the paper in \cref{subsec:notation}. Then, we will describe the computational cost of static and dynamic quantization in \cref{subsec:cost}.

% \subsection{Computational Cost} \label{subsec:cost}
The main difference between static and dynamic quantization lies in \textit{when} the quantization parameters (\cref{eq:qparams}) of the pre-activations (\ie, output) are estimated. This distinction mainly impacts the computational cost of these quantization strategies. Without loss of generality, we will analyze the working memory requirements and operation count on a generic function $f(\cdot, \vb W): \mathbb{R}^{d}\to\mathbb{R}^{h}$, parametrized by $\vb W$ and whose output entries $f_i$ can be computed independently.

% Following the notation introduced in \cref{subsec:notation}, the real-valued inference of this function is
% \begin{align} \label{eq:lin}
%     \vb y \triangleq f(\vb x, \vb W) = \vb W \vb x \; \in\mathbb{R}^{h}.
% \end{align}
For all quantization strategies, the weights are quantized before deploying the model \cite{Nagel2021AWP}, which reduces the footprint by storing the compressed weights, $Q_b(\vb W)$, and their quantization parameters. Additionally, for simplicity, let us assume the input is already quantized. This simplification is justifiable as we are only interested in modelling the computational cost induced by evaluating $f$ depending on the quantization scheme employed for the output.

Before describing the difference between static and dynamic quantization, we introduce the general algorithm employed by common neural network frameworks \cite{lai2018cmsis, david2021tensorflow} when performing operations on quantized values.
Prior to evaluating arithmetic operators, neural network inference frameworks augment the bit-width of the operands to $b' > b$\si{\bit} to avoid numeric overflow errors when representing the result. This operation is performed via data type casting denoted as $C_{b'}(\cdot)$ in the remainder of the paper. After evaluation, the result needs to be compressed back to $b$\si{\bit} based on its dynamic range. %of the result.

% After the output is computed, it is quantized with its quantization parameters to minimize memory consumption. This latter process is referred to as \emph{re-quantization} and can be a computational bottleneck depending on how the quantization parameters are estimated. 

\noindent \textbf{Static quantization.} In static quantization, all quantization parameters are estimated using a calibration dataset, as showcased in \cref{fig:quant}-a). The samples for this dataset are selected to be representative of the data distribution. The intuition behind static quantization is that if the calibration set represents the test data distribution faithfully, we can model the dynamic range of the output values of our function $f$ by evaluating it on the calibration set. This process happens \emph{before} deploying the model. Therefore, the quantization parameters of the output $(s_\text{out}, z_\text{out})$ are known before evaluating $f$. This is crucial in reducing the memory requirements of quantization, as each entry of the output vector, $\vb y_j$, can be compressed independently following
\begin{align} % \label{eq:eval-static}
    \widetilde{\vb x} &= C_{b'}(Q_b(\vb x)), \; \widetilde{\vb W} = C_{b'}(Q_b(\vb W)) \label{eq:cast} \\
    \widetilde{\vb y_j} &= f_j(\widetilde{\vb x}, \widetilde{\vb W}) \label{eq:eval_bp} \\
    \vb y_j &= Q_b\left(\widetilde{\vb y_j}, s_\text{out}, z_\text{out}\right). \label{eq:eval_q}
\end{align}

That is, we first cast to $b'$ bits the input and weight entries needed to compute the $j$-th output entry in \cref{eq:cast}. Then, we evaluate the $j$-th output entry using $b'$ bit resolution (\cref{eq:eval_bp}), and finally we quantize $\widetilde{\vb y_j}$ to $b$bit using \cref{eq:eval_q}. We emphasize that by evaluating one output entry at the time and knowing the quantization parameters a priori, we can directly quantize it to $b$ bits. Therefore, the working memory overhead of static quantization is constant for an arbitrary-size output tensor, depends on the casting bit-width ($b'$), and amounts to $3b'$\si{\bit}. Notably, there is no latency overhead, as the quantization parameters ($s_\text{out}, z_\text{out}$) are pre-computed.

\noindent \textbf{Dynamic quantization.} In dynamic quantization, the quantization parameters of all pre-activations are computed on the fly. This strategy allows for a more precise representation of the output's dynamic range, yielding improved performance. Nonetheless, this requires storing the entire output tensor $\vb y$ before computing its quantization parameters, as showcased in \cref{fig:quant}-b). Varying the size of the output tensor ($h$) scales the memory overhead of dynamic quantization linearly. Specifically,  we first need to evaluate $f$ using $b'$ bits, then extract the quantization parameters and compress $\vb y$. 
This process produces an overhead of $(b' \cdot h)$ \si{\bit}, needed to store $C_{b'}(\vb y)$. Then, we would compute its quantization parameters and quantize $\vb y$ to a $b$ bits representation. Additionally, dynamic quantization requires storing (or searching, depending on the implementation) the minimum and maximum values of the output tensor to compute the quantization parameters, incurring a minor latency overhead. % This process is summarized in \cref{alg:dynamic}.
In summary, although dynamic quantization improves robustness to input shifts, it can lead to significant working memory requirements that prevent this strategy from being applied to models for resource-constrained devices.

% Finally, we note that these conclusions do not depend on the layer used for analysis and are, therefore, valid for all layers whose output can be computed in tiles. 

\section{Methods} \label{sec:methods}
Our approach integrates the benefits of static and dynamic quantization, providing a quantization scheme with low memory overhead while adapting to each input. The core novelty of our method is the \textit{estimation} of the quantization parameters of the output tensor before executing the neural network layer, as depicted in \cref{fig:quant}-c). Notably, this combines the computational efficiency of static quantization with the flexibility of dynamic quantization. In \cref{subsec:est}, we describe the proposed estimation strategy; then, in \cref{subsec:est_cost}, we analyze its computational cost.

\begin{figure*}[h!]
    \centering
    \resizebox{\linewidth}{!}{
        \begin{tikzpicture}
            \node(img1) {
                \includegraphics[width=0.15\linewidth, trim={0 0 0cm 0}]{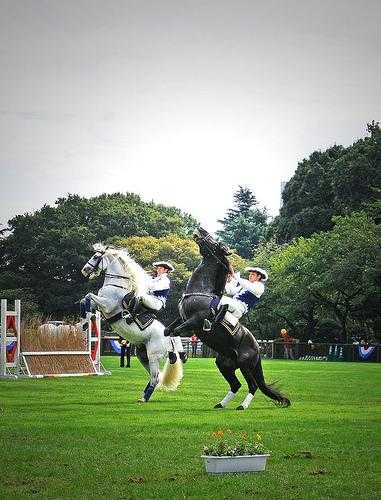}
            };
            \node[right of=img1, xshift=2cm] (img2) {
                \includegraphics[width=0.15\linewidth, trim={0 0 0cm 0}]{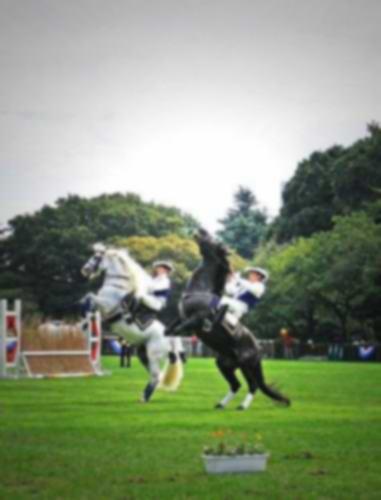}
            };
            \node[right of=img2, xshift=2cm] (img3) {
                \includegraphics[width=0.15\linewidth, trim={0 0 0cm 0}]{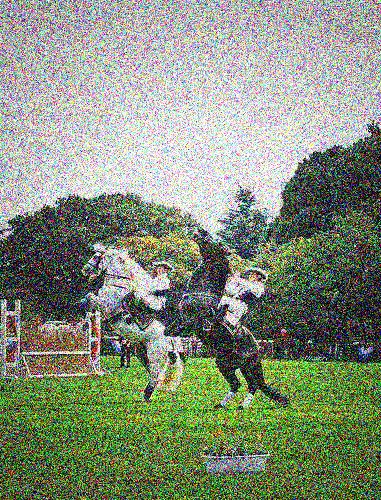}
            };
            \node[right of=img3, xshift=2cm] (img4) {
                \includegraphics[width=0.15\linewidth, trim={0 0 0cm 0}]{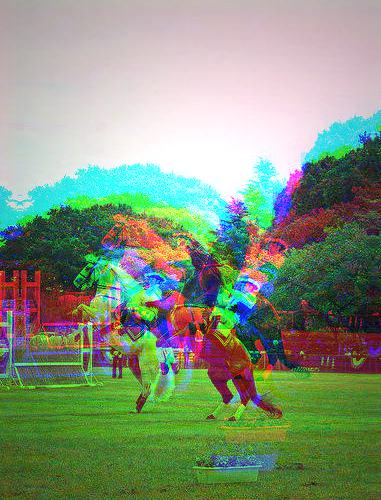}
            };
            \node[right of=img4, xshift=2cm] (img5) {
                \includegraphics[width=0.15\linewidth, trim={0 0 0cm 0}]{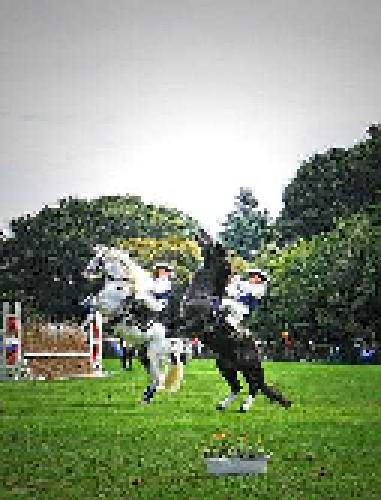}
            };
            \node[right of=img5, xshift=2cm] (img6) {
                \includegraphics[width=0.15\linewidth, trim={0 0 0cm 0}]{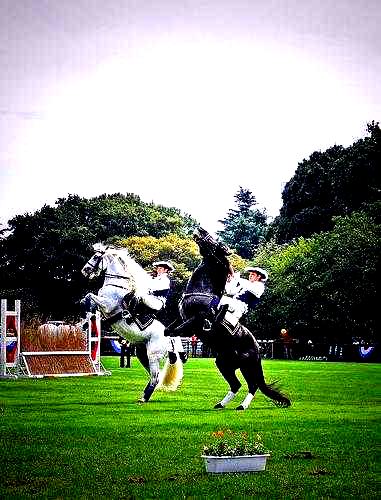}
            };
            % \node[right of=img6, xshift=2cm] (img7) {
            %     \includegraphics[width=0.15\linewidth, trim={0 0 0cm 0}]{img/corrupted_images/combined_sev3.jpg}
            % };

            % Subfigure notes
            \node [above of=img1, yshift=1cm] {\footnotesize Original image};
            \node [above of=img2, yshift=1cm] {\footnotesize Motion Blur};
            \node [above of=img3, yshift=1cm] {\footnotesize White Noise};
            \node [above of=img4, yshift=1cm] {\footnotesize Color Shift};
            \node [above of=img5, yshift=1cm] {\footnotesize Pixelate};
            \node [above of=img6, yshift=1cm] {\footnotesize Contrast};
            % \node [above of=img7, yshift=0.4cm] {\footnotesize Combinatio155};
        
        \end{tikzpicture}
    }
    \caption{This illustration showcases the effect of some of the corruptions we implemented to simulate domain shifts in the test distribution. All augmentations are shown with a severity score of three out of five.}
    \label{fig:corruptions}
\end{figure*}
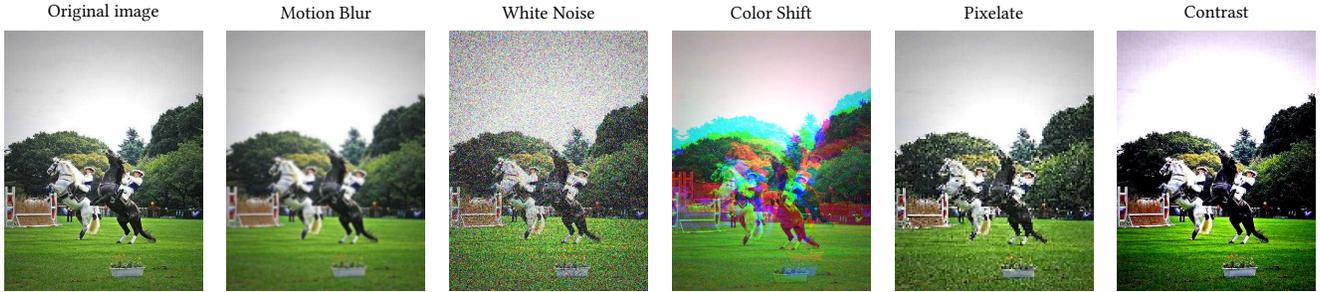

\subsection{Estimation of the Quantization Parameters} \label{subsec:est}
Our estimation strategy is based on a surrogate model of the neural network pre-activations. We designed the surrogate model exploiting an assumption commonly used when describing the propagation of pre-activations in randomly initialized neural networks \cite{lee2018deep}: neural network weights are \iid and normally distributed. The proposed model is a surrogate model since this assumption does not provably hold for trained neural networks. Therefore, we rely on an extensive empirical evaluation to verify how well the surrogate model performs on various multimedia tasks and model scales (\cref{sec:results}).
% We should note, however, that we extend this set of assumptions to trained neural networks as well. 
% Despite recent works attempting to verify these assumptions in practical use cases \cite{wolinski2022gaussian}, they refer to theoretical results originally derived for infinite-width neural networks. Additionally, their experimental setup is composed of neural network architectures with little to no computer vision applications. Therefore, we rely on an extensive empirical evaluation to verify how well these assumptions hold for various vision tasks and model scales (\cref{sec:results}).

Given this preamble, let us consider the pre-activations of a linear layer, $\vb y \triangleq\vb W\vb x$, where $\vb W\in \mathbb{R}^{h\times d}, \vb x\in\mathbb{R}^d, \vb y\in\mathbb{R}^h$ and $\vb W_{ij} \sim \mathcal{N}(\mu_W, \sigma_W^2)$. We can model the expected value of its $j$-th entry as
{\allowdisplaybreaks
\begin{align} \label{eq:exp_lin}
    \begin{split}
        \mathbb{E}[\vb y_j] &= \mathbb{E}\left[\sum_{i=1}^d \vb W_i\vb x_i \right] = 
        \sum_{i=1}^d \mathbb{E}\left[\vb W_i\right] \vb x_i =\\
        &= \mu_W \sum_{i=1}^d \vb x_i,
    \end{split}
\end{align}
}
and the variance as
{\allowdisplaybreaks
\begin{align} \label{eq:var_lin}
    \begin{split}
        \Var{\vb y_j} &= \Var{\sum_{i=1}^d \vb W_{ji}\vb x_i} = \sum_{i=1}^d \Var{\vb W_{ji}}\vb x_i^2 = \\
        &= \sigma_W^2 \sum_{i=1}^d \vb x_i^2.
    \end{split}
\end{align}
}
We observe that these results hold for all entries $j\in [1, h]$, as the \iid assumption removes all the dependencies on terms that vary with $j$. We will refer to $\Exp{\vb y_i}$ as $\Exp{\vb y}$ for brevity; a similar simplification is done for the variance. 
% Additionally, we note that in \cref{eq:exp_lin}, \cref{eq:var_lin}, we did not use the Gaussian prior assumption but only the assumption of statistical independence of the trained weights. 
Similarly, we can derive the same conclusions for a convolution. Let us consider a convolution whose kernel is $\vb K\in\mathbb{R}^{l\times p\times k\times k'}$, where $\vb K_{vrij} \sim \mathcal{N}(\mu_{K,v}, \sigma_{K,v}^2)$, $l$ and $p$ represent the output and input channels, respectively. We can easily derive the expected value of the entries $\vb y_{ijv}$ as
{\allowdisplaybreaks
\begin{align} \label{eq:exp_conv}
    \begin{split}
        \Exp{\vb y_{ijv}} &= 
        % \Exp{\sum_{r=1}^p \sum_{q=-\frac{k}{2}}^\frac{k}{2} \sum_{t=-\frac{k'}{2}}^\frac{k'}{2} \vb K_{qtvr} \vb x_{(i+q)(j+t)r}} = \\
        \mu_{K,v} \sum_{r=1}^p \sum_{q=-k/2}^{k/2} \sum_{t=-{k'/2}}^{k'/2} \vb x_{(i+q)(j+t)r},
    \end{split}
\end{align}
}
and their variance as
{\allowdisplaybreaks
\begin{align} \label{eq:var_conv}
    \begin{split}
        \Var{\vb y_{ijv}} &= 
        % \Var{\sum_{r=1}^p \sum_{q=-\frac{k}{2}}^\frac{k}{2} \sum_{t=-\frac{k'}{2}}^\frac{k'}{2} \vb K_{qtvr} \vb x_{(i+q)(j+t)r}} = \\
        \sigma^2_{K,v} \sum_{r=1}^p \sum_{q=-k/2}^{k/2} \sum_{t=-{k'/2}}^{k'/2} \vb x^2_{(i+q)(j+t)r}.
    \end{split}
\end{align}
}
% These results follow by observing that each entry in the output tensor $\vb y_{ij}\in\mathbb{R}^l$ is the sum of the output of the $k\times k'$ linear layers applied on the input patch. 
Contrary to what we observed in \cref{eq:exp_lin,eq:var_conv}, we note that, in this case, each output entry's expected value and variance depend on the $i,j$ and $v$ indices. This dependence on $i,j$ comes from the shifting of the kernel over the convolution's input, which causes the input to change. The dependence on $v$ derives from modelling the mean and variance per channel. Assuming that $\sigma_{K,v}^2=\sigma_{K,v'}^2$ for all channels $(v, v') \in [1,p]^2$ removes the channel dependency and can be appropriate depending on the quantization scheme.

In our framework, we aggregate the prediction of \cref{eq:exp_conv,eq:var_conv} computed for each output using the following
{\allowdisplaybreaks
\begin{align} \label{eq:aggr}
    \begin{split}
        \Exp{\vb y}&=\frac{1}{HWp}\sum_{v=i=j=1}^{p,H,W} \Exp{\vb y_{ijv}}, \\
        \Var{\vb y}&=\sum_{v=i=j=1}^{p,H,W} \Var{\vb y_{ijv}}^2 + (\Exp{\vb y_{ijv}} - \Exp{\vb y})^2,
    \end{split}
\end{align}
}
where $\sum_{v=i=j=1}^{p,H,W}$ is a shorter notation for $\sum_{v=1}^p\sum_{i=1}^H\sum_{j=1}^W$ and $H,W$ are the output's height and width. This allows us to model the behavior of the output activations on a per-tensor or per-channel resolution, without requiring per-pixel quantization parameters. 
% average value of all means variances to aggregate the results over the spatial axis. Therefore, in this case, we simplify the notation using $\Exp{\vb y}$ and $\Var{\vb y}$ to indicate the aggregated estimate. 

The remaining step is to extract the quantization parameters from the estimates of $\Exp{\vb y}$ and $\Var{\vb y}$. To do so, we compute the probability of each layer's pre-activations being within an asymmetric interval $I(\alpha,\beta) = [\mu_{\vb y} - \alpha\sigma_{\vb y}, \mu_{\vb y} + \beta\sigma_{\vb y}]$ as
\begin{align}
    \P{y \in I} = \frac{1}{\#\mathcal{S}}\sum_{y_i\in\mathcal{S}} \indicator_{y_i \in I},
\end{align}
where $\mathcal{S}$ is a calibration dataset of $\#\mathcal{S}$ elements, $\indicator_\cdot$ is the indicator function, and $\mu_{\vb y} \triangleq \Exp{\vb y}, \sigma^2_{\vb y} \triangleq \Var{\vb y}$. We use this to estimate the coverage of the layer's pre-activations given $I(\alpha, \beta)$ and tune the $\alpha,\beta$ hyperparameters to represent a given percentage of the pre-activations. We emphasize that $\alpha,\beta$ remain fixed after calibration. Finally, we extract the quantization parameters using $I(\alpha,\beta)$ as the pre-activations' dynamic range. The design of $I(\alpha,\beta)$ scales $s_\text{out}$ proportionally to the pre-activations' dispersion, while $z_\text{out}$ depends linearly on the mean of the distribution. Therefore, our method is a trade-off between static and dynamic quantization: we need a calibration dataset, as per static quantization, but our quantization scheme adapts the quantization parameters based on each input.

\subsection{Computational Overhead of the Estimation} \label{subsec:est_cost}
This section presents the computational overhead of our estimation strategy for linear layers and convolutions. We note that both layers are part of the set of functions described in \cref{sec:model_q}. Therefore, these results are comparable with those of \cref{sec:model_q}. Note that our estimation strategy runs before evaluating the function $f$, as shown in \cref{fig:quant}. Therefore, the computational overhead of our quantization scheme equals that of static quantization, plus the cost of estimating the quantization parameters from the input and weights.

% Therefore, we want to (i) ensure that the working memory requirement of the parameter estimation is lower than that of dynamic quantization and (ii) design a scaling strategy that allows the computational overhead of the model to be adapted. 

For both linear layers and convolutions, the memory overhead of the parameter estimation is constant and equal to $2b'$\si{\bit}, equally divided between the mean and the variance estimation. This overhead adds a marginal cost to the memory requirement of static quantization while drastically reducing the memory requirement of dynamic quantization.

The time complexity differs depending on the layer. For linear layers, the time complexity of evaluating \cref{eq:exp_lin,eq:var_lin} is the same and scales as $\mathcal{O}(d)$, where $d$ is the number of input channels. Once $\Exp{\vb y}$ and $\Var{\vb y}$ are estimated for one entry, they are known for the entire output vector. Therefore, the linear layer's computational overhead is constant with respect to the output tensor's shape. 
For the convolution, instead, the time complexity of evaluating \cref{eq:exp_conv,eq:var_conv} for one entry of the output tensor scales as $\mathcal{O}(pkk')$, where $p$ is the number of input channels and $k,k'$ represent the kernel shapes. Assuming that the output tensor has resolution $H\times W$, the overall cost scales as $\mathcal{O}(HWlpkk')$, as per the corresponding convolution. This can impact the performance of processing in real-time high-resolution images. Nonetheless, we emphasize that, with our strategy, we can trade working memory for inference time. This is an asset for deploying neural networks on resource-constrained devices such as microcontrollers, where working memory is a hard constraint.

To mitigate the computational overhead of the estimation stage, we introduce a scaling parameter called \emph{sampling stride}, $\gamma \in \mathbb{R} : 0 < \gamma \le \max{(H,W)}$. This hyperparameter expresses the number of output patches $\vb y_{ikv}$ for which the estimation is computed as a fraction of the output resolution. Therefore, it scales the complexity of the estimation stage quadratically, rendering the complexity $\mathcal{O}(HWlpkk'\gamma^{-2})$. The scaling induced by the sampling stride is studied in \cref{subsec:res_comp}, and its impact on downstream task performance in \cref{subsec:res_perf}.
\section{Experimental design} \label{sec:expdesign}
% To validate the proposed strategy on various vision tasks, we designed an experimental setup composed of five vision tasks: object detection, object segmentation, pose estimation, and oriented bounding box prediction. 
To validate the method, we designed an experimental setup that comprises three stages. First, we analyze the computational cost of the proposed quantization approach when deployed on a microcontroller (\cref{subsec:hardware}). Then, we evaluate the performance with respect to the static and dynamic quantization baselines across several computer vision tasks, i.e. object detection, object segmentation, pose estimation, oriented bounding box prediction and image classification (\cref{subsec:moddata}). And finally, we conduct a sensitivity study to evaluate the impact of the calibration's set size and complexity reduction hyperparameters (\cref{subsec:sensitivity}). 
% baselines In this section, we present the datasets and models employed to validate the model's performance. Finally, we present the hardware platform used for on-device computational cost analysis.

\begin{figure*}[t]
    \centering
    \resizebox{\linewidth}{!}{
        \begin{tikzpicture}
            \node(img1) {
                \includegraphics[width=0.3\linewidth]{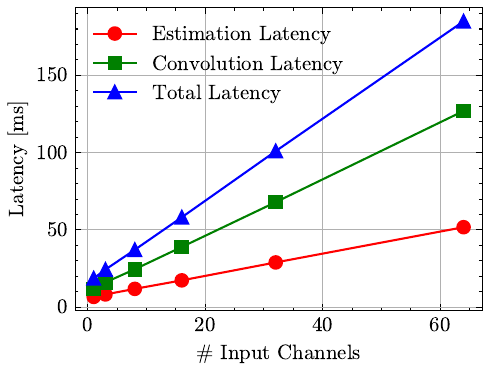}
            };
            \node[right of=img1, xshift=4.5cm] (img2) {
                \includegraphics[width=0.3\linewidth]{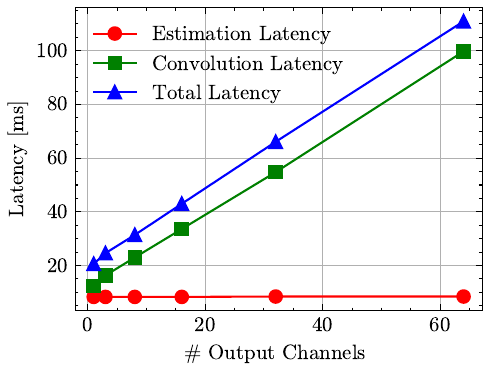}
            };
            \node[right of=img2, xshift=4.5cm] (img3) {
                \includegraphics[width=0.29\linewidth]{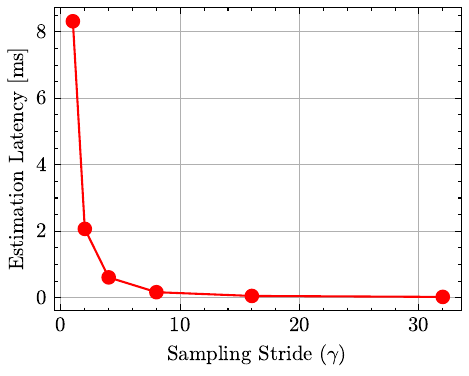}
            };

            % Subfigure notes
            \node [below of=img1, yshift=-1.5cm] {a)};
            \node [below of=img2, yshift=-1.5cm] {b)};
            \node [below of=img3, yshift=-1.5cm] {c)};
        \end{tikzpicture}
    }
    \vspace{-20pt}
    \caption{
        This diagram showcases the latency analysis of convolution. We vary independently the number of input and output channels. From left to right, the hyperparameters for the latency measurements were a) input shape 32x32x$C_\text{in}$, output channels 3, convolution's stride 1, b) input shape 32x32x3, convolution's stride 1, and c) input shape 32x32x3, convolution's stride 1.
}
    \label{fig:results_comp}
\end{figure*}
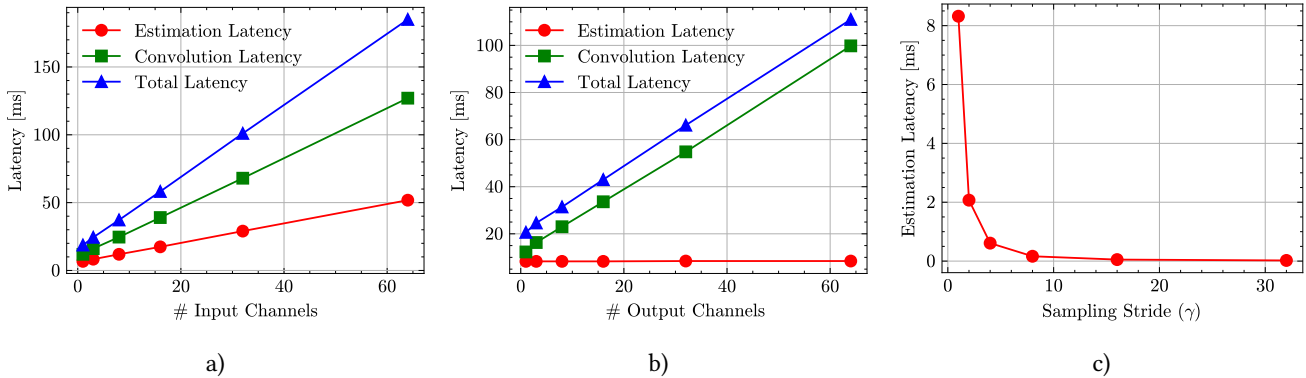

\subsection{On-Device Complexity Analysis} \label{subsec:hardware}
To evaluate the latency of our approach on real-world embedded hardware, we integrated our method into CMSIS-NN\cite{lai2018cmsis}—a widely adopted neural network inference library for microcontrollers. We created a wrapper around the CMSIS-NN functions \texttt{arm\_convolve\_s8} and \texttt{arm\_fully\_connected\_s8}, which are designed for int8 convolution and linear operations, respectively. All computations were carried out using fixed-point arithmetic to ensure full hardware compatibility. The square root function, needed for estimating the standard deviation, is computed using the Newton-Raphson algorithm \cite{newton}.

We conducted latency measurements on an STM32L476RG development board using our release of the CMSIS library. To perform accurate measurements, we toggled a GPIO pin after each inference, and used an oscilloscope to record the timing data. The reported latency values represent the averages computed over 100 runs.

\subsection{Datasets and Models} \label{subsec:moddata}
\noindent \textbf{Datasets.} For object detection, we used the Common Objects in Context (COCO) dataset \cite{Lin2014MicrosoftCC}. COCO contains 330K images among 80 object categories, with 200K images also annotated for pose estimation and semantic segmentation. For pose estimation and semantic segmentation, we evaluated our method on the COCO dataset. Finally, for oriented bounding boxes (OBB), we used the DOTAv1 data \cite{Xia_2018_CVPR}. DOTAv1 is a large-scale benchmark for object detection from aerial images. OBB aims to enclose the target objects inside the bounding box exactly. 
For image classification, we considered the ImageNet1k \cite{5206848} benchmark.

To study the impact of domain shifts on the proposed quantization scheme and the baselines, we conducted an out-of-domain evaluation of the model. We benchmarked the pretrained models on images corrupted with randomly sampled augmentations: white noise injection, blurring, pixelation, quantization, color shift, brightness changes and contrast. For each augmentation, we defined a severity score from one to five. When using a severity of five, the image is still recognizable by the human eye\footnote{We invite the reader to refer to the code to learn about the specific values used for each augmentation.}. We evaluate the models by uniformly sampling an augmentation and severity for each image. On top of the set of augmentations described above, we also added a `combination' option that would combine more augmentations in one single inference. A sample of the out-of-distribution samples is presented in \cref{fig:corruptions}.

\noindent \textbf{Models.} To evaluate our quantization scheme, we used the YOLO11 models \cite{yolo11_ultralytics} for all tasks excluding image classification, specifically utilizing the N variant with official Ultralytics checkpoints across all datasets.  For image classification, we use two widely adopted convolutional neural networks, namely ResNet50 \cite{he2016deep} and MobileNetv2 \cite{sandler2018mobilenetv2}. For both models, we used pretrained checkpoints provided by the \texttt{torchvision.models} library.

For all models, we tested both per-channel and per-tensor quantization. The calibration set for our approach and static quantization is shared and composed of 16 images from the training sets of each dataset. During the evaluation of the performance, we emulate the quantization pipeline using a custom-made quantization API, with a fixed bitwidth of \SI{8}{\bit}. All tensors and layers are treated in the same way for all quantization approaches (\eg, layer folding is never applied).

\subsection{Sensitivity study} \label{subsec:sensitivity}
The aim of our sensitivity study is twofold: (i) exploring how the complexity reduction from the sampling stride impacts performance and (ii) understanding the impact of the calibration set's size on performance. We consider the ResNet50 checkpoint evaluated on ImageNet. We experimented with sampling strides $\gamma\in\{1, 4, 8, 16, 32\}$. This reduces quadratically the complexity of the estimation procedure. Then, in the interest of time, we selected the model with the best sampling stride ($\gamma=4$) for our ablation on the impact of the calibration set size. We evaluated the proposed quantization scheme with three sets for each explored cardinality $\#\mathcal{S}=\{16, 32, 64, 128, 256, 512\}$ to mitigate the impact of selecting more representative dataset samples.

\section{Results} \label{sec:results}

In this Section, we present the experimental results. \cref{subsec:res_comp} presents the latency analysis. \cref{subsec:res_perf} reports the performance of our method compared to the baselines. Finally, \cref{subsec:res_sens} describes the sensitivity analysis results.

\subsection{Computational complexity} \label{subsec:res_comp}
In \cref{fig:results_comp}, we showcase in green the overall latency to perform a convolution quantized per tensor using our quantization strategy in blue. The latency of the estimation and convolution is reported in red and green, respectively. We analyzed the scaling behavior of the latency with respect to the number of input and output channels. We varied one parameter at a time to characterize the impact of different convolution hyperparameters.
As expected by analyzing the analytical complexity model in \cref{subsec:est_cost}, we observe that the latency is linearly increasing with the number of channels. At the same time, it remains constant independently of the number of output channels. In \cref{fig:results_comp}-c), we showcase how the estimation latency is affected by different values of sampling stride. Again, we verify the quadratic decrease in latency modeled in \cref{subsec:est_cost}. We emphasize that, while latency increases linearly with the number of input channels, it decreases quadratically with sampling stride. Therefore, we can tune the sampling stride parameter to compensate for increasing latency costs without compromising performance significantly (see \cref{subsec:res_sens}). Overall, on-device experiments on an STM32L476RG development board showcase that the proposed quantization approach enables dynamic quantization with negligible memory overhead and a minimal, tunable computation overhead.

\subsection{Performance Evaluation} \label{subsec:res_perf}
\noindent \textbf{In-Domain.} In \cref{tab:res_id}, we present the performance of the proposed quantization scheme and the baselines on in-domain data. Consistently among tasks, dynamic quantization performs best, with a minor performance degradation with respect to the full precision model. The average performance degradation over all tasks is 1.09\% mAP$_\text{50-95}$ for per-tensor quantization and 0.54\% mAP$_\text{50-95}$ for per-channel quantization. Our method is always the second-best, with an average performance degradation of 1.64\% mAP$_\text{50-95}$ for per-tensor quantization and 0.88\% mAP$_\text{50-95}$ for per-channel quantization. We emphasize that, due to its memory requirements, dynamic quantization is not applicable in some application domains. On the contrary, our quantization strategy yields comparable results with only a fraction of the required memory, thus enabling on-device dynamically-quantized inference. Compared to static quantization, our approach yields better performance, especially in the per-channel setting, where the mAP difference can be up to 8\% (ResNet50 on ImageNet1K). Our approach is a good tradeoff between static and dynamic quantization performance, with a computational cost between these two baselines.

\begin{table*}[]

    \centering
    \begin{tabular}{lllccc|cc|cc}
        \toprule
        % \cline{5-10}
        \multicolumn{1}{c}{} & \multicolumn{1}{c}{} & \multicolumn{1}{c}{} & \textbf{} & \multicolumn{2}{|c|}{\textbf{Ours}} & \multicolumn{2}{c|}{\textbf{Dynamic}} & \multicolumn{2}{c}{\textbf{Static}} \\ \midrule
        \multicolumn{1}{l}{\textbf{Task}} & \multicolumn{1}{l}{\textbf{Dataset}} & \multicolumn{1}{l|}{\textbf{Model}} & \multicolumn{1}{c|}{\textbf{FP32}} & {\textbf{T}} & {\textbf{C}} & {\textbf{T}} & {\textbf{C}} & {\textbf{T}} & {\textbf{C}} \\ \midrule
        Detection & COCO & \multicolumn{1}{l|}{Yolo11n-det} & \multicolumn{1}{c|}{0.3923} & \textit{0.3889} & \textit{0.3883} & \textbf{0.3901} & \textbf{0.3907} & 0.3877 & 0.3401 \\
        Segment & COCO & \multicolumn{1}{l|}{Yolo11n-seg} & \multicolumn{1}{c|}{0.3204} & \textit{0.3155} & \textit{0.3157} & \textbf{0.3159} & \textbf{0.3176} & 0.3152 & 0.3013 \\
        Pose & COCO-POSE & \multicolumn{1}{l|}{Yolo11n-pose} & \multicolumn{1}{c|}{0.5114} & \textit{0.5083} & \textit{0.5101} & \textbf{0.5089} & \textbf{0.5111} & 0.4971 & 0.2135 \\
        OBB & DOTAv1 & \multicolumn{1}{l|}{Yolo11n-obb} & \multicolumn{1}{c|}{0.4706} & \textit{0.4672} & \textit{0.4688} & \textbf{0.4688} & \textbf{0.4705} & 0.4536 & 0.4305 \\
        Classification & ImageNet1K & \multicolumn{1}{l|}{Resnet50} & \multicolumn{1}{c|}{80.858} & \textit{0.7764} & \textit{0.7800} & \textbf{0.7916} & \textbf{0.8026} & 0.7797 & 0.7124 \\
        Classification & ImageNet1K & \multicolumn{1}{l|}{MobileNetv2} & \multicolumn{1}{c|}{72.022} & \textit{0.7141} & \textit{0.7131} & \textbf{0.7142} & \textbf{0.7154} & 0.7123 & 0.6826 \\ \bottomrule
    \end{tabular}
    \vspace{5pt}
    \caption{Performance comparison across datasets and tasks for In-Domain samples. For detection, pose estimation, segmentation, and OBB, the numbers refer to mean Average Precision (mAP$_\text{50-95}$). For image classification, we report top-1 accuracy. T and C refer to per-tensor and per-channel quantization, respectively. Bold numbers represent the best performance per row; italic numbers represent the second best. For our approach, we used $\gamma=1$.}
    \label{tab:res_id}

\end{table*}
\begin{table*}[]
    \centering
    \begin{tabular}{lllc|cc|cc|cc}
        \toprule
        \multicolumn{1}{c}{} & \multicolumn{1}{c}{} & \multicolumn{1}{c}{} & \textbf{} & \multicolumn{2}{c|}{\textbf{Ours}} & \multicolumn{2}{c|}{\textbf{Dynamic}} & \multicolumn{2}{c}{\textbf{Static}} \\ \midrule
        \textbf{Task} & \textbf{Dataset} & \multicolumn{1}{l|}{\textbf{Model}} & \textbf{FP32} & \textbf{T} & \textbf{C} & \textbf{T} & \textbf{C} & \textbf{T} & \textbf{C} \\ \midrule
        Detection & COCO & \multicolumn{1}{l|}{Yolo11n-det} & 0.2167 & \textit{0.1934} & \textit{0.1922} & \textbf{0.2099} & \textbf{0.2108} & 0.1917 & 0.1755 \\
        Segment & COCO & \multicolumn{1}{l|}{Yolo11n-seg} & 0.1679 & \textbf{0.1723} & \textbf{0.1728} & {0.1624} & \textit{\textit{0.164}} & \textit{0.1718} & 0.1581 \\
        Pose & COCO-POSE & \multicolumn{1}{l|}{Yolo11n-pose} & 0.2884 & \textit{0.2822} & \textit{0.2809} & \textbf{0.2850} & \textbf{0.2873} & 0.2135 & 0.2087 \\
        OBB & DOTAv1 & \multicolumn{1}{l|}{Yolo11n-obb} & 0.2524 & \textit{0.2434} & \textit{0.2454} & \textbf{0.2500} & \textbf{0.2504} & 0.2326 & 0.2267 \\
        Classification & ImageNet1K & \multicolumn{1}{l|}{Resnet50} & 0.4963 & \textit{0.4641} & \textit{0.4918} & \textbf{0.4831} & \textbf{0.4961} & 0.4583 & 0.3879 \\
        Classification & ImageNet1K & \multicolumn{1}{l|}{MobileNetv2} & 0.3867 & \textit{0.3829} & \textit{0.3838} & \textbf{0.3856} & \textbf{0.3876} & 0.3803 & 0.3523 \\ \bottomrule
    \end{tabular}
    
    \vspace{10pt}
    \caption{Performance comparison across datasets and tasks for Out-of-Domain samples (\ie, with corruptions). For detection, pose estimation, segmentation, and OBB, the numbers refer to mean Average Precision (mAP$_\text{50-95}$). For image classification, we report top-1 accuracy. T and C refer to per-tensor and per-channel quantization, respectively. Bold numbers represent the best performance per row; italic numbers represent the second best. For our approach, we used $\gamma=1$.}
    
    \label{tab:res_ood}
\end{table*}
\begin{figure}[t]
    \centering
    \includegraphics[width=0.87\linewidth]{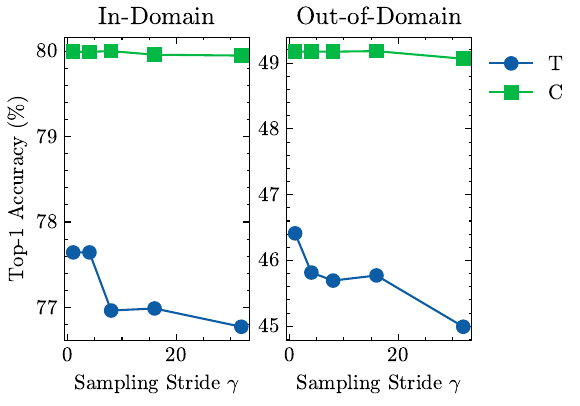}
    \caption{Impact of sampling stride, $\gamma$, on per-tensor (T) and per-channel (C) quantization.}
    \label{fig:sampling_stride}
\end{figure}

\noindent \textbf{Out-of-Domain.} In \cref{tab:res_ood}, we present the performance of the proposed quantization scheme and the baselines on out-of-domain data. This evaluation is designed to stress-test the quantization strategies. The intuition is that ours and static quantization would be the most challenged by the mismatch between the calibration and test samples. We observe a similar trend to that of the In-Domain setting. Notably, the performance across all tasks, models and bit resolutions dropped significantly due to the additional complexity of performing these tasks on noisy images. Nonetheless, dynamic quantization remains the overall best approach, with an average performance drop of 1.00\% per-tensor and 0.18\% per-channel. We note that our strategy is the best for segmentation, with static and dynamic performing within 1\% mAP. Similarly to what we observed during the In-Domain evaluation, our strategy strikes the optimal tradeoff between complexity and performance. Its performance, averaged across all benchmarks, is 2.58\% per-tensor and 0.99\% per-channel, while static achieves 5.62\% per-tensor and 12.6\% per-channel, respectively.

Another trend that emerges from the data for both the In-Domain and Out-of-Domain evaluations is that per-channel quantization performs better than per-tensor quantization for ours and dynamic quantization. For static quantization, instead, per-tensor quantization works better. One possible explanation is that when performing static quantization per-tensor, the inaccuracies in the calibration of the quantization parameters for a specific channel can be `hidden' by the parameter selection of the other channels. During per-channel calibration, a wrong configuration of the quantization parameters almost certainly results in a performance drop.

\subsection{Sensitivity study} \label{subsec:res_sens}

\begin{figure}[t]
    \centering
    \includegraphics[width=0.815\linewidth]{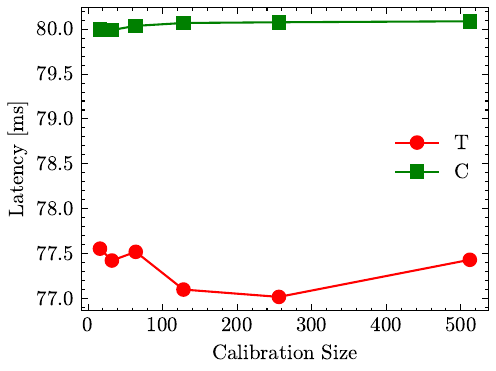}
    \caption{Impact of calibration dataset size, $\#\mathcal{S}$ on per-tensor (T) and per-channel (C) quantization.}
    \label{fig:calib}
\end{figure}

In \cref{fig:sampling_stride}, we showcase how varying the sampling stride impacts the performance of our quantization strategy. The per-channel quantization approach maintains almost constant performance with increasing $\gamma$ for both the In-Domain and Out-of-Domain settings. This suggests that per-channel quantization is robust to spatial shifts in the latent representations. For per-tensor quantization, instead, the performance of the quantized model drops by $\sim$2\% when reducing the complexity of the estimate by a factor of 1024. We note that this performance drop is negligible considering the significant benefits in latency. Therefore, from these results and those presented in \cref{subsec:res_comp}, we conclude that our quantization strategy can be adapted to adjust the complexity of the estimation, without sacrificing performance significantly.

Finally, we analyzed the impact of the calibration set size on the performance of our approach. We report the results in \cref{fig:calib}. For per-channel quantization, we achieved good performance with only 16 images and observed negligible improvements with larger calibration datasets. For per-tensor calibration, there is more variability in performance with respect to the size of the calibration set. Nonetheless, the variability is lower than 1\% in top-1 accuracy with no evident relationship with the size of the calibration set. We conclude that the calibration set size does not play an essential role in quantization performance in our experimental setup.

\section{Conclusion} \label{sec:conclusion}
%POSSIBILE TEMPLATE DI CONCLUSIONE... ma cancellate se non ha senso.
This work introduced a probabilistic framework for dynamic quantization that bridges the gap between static and dynamic quantization. Our method leverages a surrogate model to estimate the dynamic range of pre-activations before layer execution, thus enabling input-adaptive quantization with a memory overhead in the same order of static quantization.
We validated our approach across diverse vision tasks and architectures, demonstrating that it performs comparably to dynamic quantization, while maintaining a latency and memory profile comparable to static quantization. Furthermore, we showed that our method is robust to distribution shifts and can be tuned efficiently via the sampling stride parameter, enabling deployment on resource-constrained devices such as microcontrollers.
Our findings suggest modeling pre-activation via surrogate statistics is a viable and scalable alternative to conventional dynamic quantization for neural networks.

% ACKNOWLEDGEMENT...  non li mettiamo ora, ma andrebbero messi a FAIR, a IPCEI, eccetera...
%This work was supported by Ministero delle Imprese e del Made in Italy (IPCEI Cloud DM 27 giugno 2022 – IPCEI-CL-0000007) and European Union (Next Generation EU).

\section{Acknowledgements}
This work was supported by Ministero delle Imprese e del Made in Italy (IPCEI Cloud DM 27 giugno 2022 – IPCEI-CL-0000007) and European Union (Next Generation EU). We acknowledge ISCRA for awarding this project access to the LEONARDO supercomputer, owned by the EuroHPC Joint Undertaking, hosted by CINECA (Italy).

\bibliographystyle{ACM-Reference-Format}
\bibliography{main.bib}

\end{document}